\documentstyle[12pt]{article}

\author{
Jos\'{e} M. Go\~{n}i \\
Jos\'{e} C. Gonz\'{a}lez \\
{\em E.T.S.I. Telecomunicaci\'{o}n} \\
{\em Universidad Polit\'{e}cnica de Madrid} \\
{\em 28040 Madrid, SPAIN} \\
{\em Phone: +34 1 336.72.87}\\
{\tt E-mail: jmg@mat.upm.es} \\
}

\title{
A framework for lexical representation
}

\date{}

\begin{document}

\pagestyle{empty}

\maketitle

\begin{abstract}
In this paper we present a unification-based lexical platform designed for
highly inflected languages (like Roman ones). A formalism is proposed for
encoding a lemma-based lexical source, well suited for linguistic
generalizations. From this source, we automatically generate an allomorph
indexed dictionary, adequate for efficient processing.  A set of software
tools have been implemented around this formalism: access libraries,
morphological processors, etc.
\end{abstract}

\thispagestyle{empty}

\section{Motivation and features}

The lexical framework presented here was born from the necessity of
representing lexical information for the ARIES\footnote{ARIES is a project
funded by the Spanish National Programme of R+D entitled: {\em An
Architecture for Natural Language Interfaces with User Modeling}.} project
\cite{Go93}, where it is integrated. We are going to describe here the
language used to represent surface aspects of the lexicon, the devices
designed, and the methods to develop application oriented, fast access
dictionaries.

We summarize here some features of our representation language:

{ \bf Expressiveness: } All the information needed in the lexical database can
be expressed, and structure can be imposed on it. Linguistic generalizations
are captured by grouping related entries in {\em lemmas}, or by using the
mechanism of  information inheritance.

The information related to a lemma is structured in a tree-shaped feature
bundle attached to it. This tree structure is, we think, powerful enough to
represent the relevant information, so the more general structure of a directed
acyclic graph was discarded. This decision has proved to be right for
morphology related information or low level syntactic features.

{\bf Versatility: } Different implemented  applications  have different
lexical interfaces, that are designed in a programming language dependent way.
In our approach translation to other representation formats and languages is
easily done in a non-ambiguous way.

{\bf Economy of expression: } The syntactic overhead needed to structure the
information has been reduced to a minimum, without compromising neither the
expressive ability nor the non-ambiguity of the syntax of the formalism.

This feature is in permanent conflict with readability, although the latter is
not strongly degraded, since source lexical databases are intended to be
liable to edition by hand with any text editor\footnote{User-friendly tools
are being designed to help in this process to a human lexicographer.}.

{\bf Non redundancy: } Redundant information is kept to a minimum by
exploiting the default inheritance devices and the notational abbreviations
included, such as value disjunction.

\section{Influence of the Spanish Morphology in the Design}

The Spanish language strongly relies on morphology for word formation.  This
is particularly true for verbs, which have a different word form for each
different combination of mood, tense, person and number. For that reason a
majority of Spanish verbs have as much as 53 different simple (word) forms.

Nouns and adjectives (nominals) have also different forms, depending of
the combination chosen of gender and number, so a maximum of four different
word forms are needed for these part-of-speech {\em lemmas}.

So, for any serious natural language processing application built for Spanish,
an account of morphology is needed, or at least of inflectional
morphology\footnote{The treatment of compositional or derivative morphology is
not as critical, and, to our knowledge, it does not exists any formalized
theoretical linguistic description of such phenomena.}. This is true not only
for reducing the size of the lexicon to a manageable level, but also for
capturing the linguistic fact that different entries (word forms) are strongly
related.

For the treatment of inflectional morphology for Spanish verbs, nouns and
adjectives we use the model developed in \cite{Mor91}. Such treatment is also
described in \cite{MorGo94}. We summarize here its main features:

\begin{itemize}

\item Morphological processing is constrained to morpheme concatenation, so
its allomorphic variants have to be stored or computed.

\item The model follows a {\em Graphical Word criterion}, that only considers
relevant its written form. This criterion requires that additional allomorphs
be necessary in some cases, because of diacritical marks, or different surface
realization of the same phoneme in different contexts.

\item Feature unification is the information combining device used to select
the relevant allomorphic variants to be concatenated. Allomorphs have an
attached feature structure DAG, and two or more of them are concatenated only
if their feature structures are validated by context-free word formation
rules\footnote{Rules currently used are like PATR-II ones described in
\cite{Shi86}.}.

\item Models for verbs and nominals are described in order to capture some
interesting and well founded linguistic generalizations. These models capture
regularities in the inflectional behavior of the Spanish verbs and nouns.

\item Some lexicalized forms are included for very irregular word forms that
can not be included in any of the proposed models.

\end{itemize}

Some changes are introduced in this approach, just to improve some of the
ideas above: First, we merge all the allomorphic variants for a
particular verb or nominal in the same entry or {\em lemma\/}; and second, we
introduce inheritance to capture generalizations that could be missed
otherwise. These generalizations include general rules permitting to compute
allomorphic variants of a {\em lemma\/} for a given model, in a similar way as
in \cite{Haus91}. Inheritance can be prohibited in a particular feature for
selected irregular or special entries, just by assigning a value to that
feature.

This design leads towards  considering two different lexical levels:
\begin{description}

\item[Source Lexical Base:] This level captures linguistic
generalizations by merging related allomorph entries, by considering classes
of {\em lemmas\/} or by specifying rules to compute different allomorphic
variants. Inflectional morphemes --that constitute a closed group-- are also
included, as well as the set of lexicalized word forms.  This level concerns
to agents\footnote{These agents can be human (lexicographers) or a learning
computer process.} editing the database.

\item[Object Dictionary:] This level is related to the computer processing of
the lexical knowledge included in the Lexical Base. To facilitate such
process, the lexical entries are expanded to different allomorphs that
constitute the key entries in this level. It is automatically derived from the
Source Lexical Base.

\end{description}

An entry, at any of the levels, is composed of a name or label, that
constitutes the access key for that entry, and an attached {\em feature
structure\/}.  We use the term EN ({\em Entry Name\/}) for the label and ES
({\em Entry Structure\/}) for the attached feature structure.

Most entries in the Source Lexical Base are {\em lemmas\/} grouping related
word forms. As each one can have different surface realizations, the relevant
allomorphs are included as values of particular features in the ES attached to
the label\footnote{We use the infinitive form as the label of a particular
verb, and the singular masculine/feminine for nouns.}.

In the Object Dictionary, the roles of lemma identifiers and allomorphs are
different, since for each entry, the EN is one of the allomorphic variants, and
the lemma label is kept as the value of a particular feature included in its
attached ES.

This paper is concerned with the representation language selected for the
Source Lexical Level, that also includes rules for expressing how the
mapping  to the Object Dictionary will be achieved. The representational
issues for the latter are considered implementation details and will not be
considered here\footnote{Intensive development related with this subject is
being carried at our site.}.

\section{The Language}

The main characteristics of the Source Lexical Base representation language
are described in this section. Further extensive description can be found in
\cite{Go93}. Examples will be given as needed, in order to illustrate the use
of the language and its semantics, and special encoding conventions, since a
formal description will not be presented.

As we stated above, each entry in this Source Lexical Base, is composed of an
EN, or label, and a ES, or feature structure, restricted to be a tree. The ES
has a number of labeled features, that can have an atomic value --a label
assigned to that feature--, or a structured one --another feature
structure. As a particular case, a string of characters can be encoded as an
atomic value for a feature. Value assignment to a feature is achieved by an
equation in the form:
\[
p = v_1\; v_2\; \ldots \; v_n
\]
where $p$ is a sequence of one or more blank space separated labels that
constitute a path for accessing the feature from the root of the tree. The
$v_i$ are the atomic values that this particular feature can have. Only one
value is permitted if it is of character string type. Paths in the left hand
side of the equations are the mechanism provided to define a tree-shaped
feature structure, and the multiple-valued features are provided as a
notational shorthand for disjunction.

The Source Lexical Base is split into sections, each one headed by a special
keyword. An include facility is also provided in order to promote physical
division of the Lexical Base into different computer files. The sections that
can appear in the Source Lexical Base are reviewed below.

\subsection{Morphemes and words}

The morphemes section is intended for the inclusion of inflectional morphemes
with a grammatical function. These morphemes usually convey grammatical
information such mood, aspect, person and tense (verbs), or gender and number
(nouns or adjectives). The entries in this section will pass almost unchanged
to the Object Dictionary upon compilation. One example is provided for the
verbal ending {\em -\'{abamos}}, with agreement, tense and mood features, as
well as a concatenating category ({\tt concat}) imposed by the morphological
rules, and two features ({\tt stt, sut}) that restrict possible concatenations
for that morpheme:
\begin{quote}\begin{verbatim}
#MORPHEMES

'abamos
agr pers = 1
agr num = plu
vinfo tense = impf
vinfo mood = ind
conj = 1
stt = 24
sut = reg
concat = vm
\end{verbatim}\end{quote}

The words section is intended for lexicalized words that are included {\em as
is\/} in the Source Lexical Base. The more frequent {\em clients\/} of this
section are very irregular words, usually with an auxiliary function. The
entries in this section will also  pass almost unchanged to the
Object Dictionary. The section is provided to physically separate morphemes
from words, although the behavior of the entries in both sections will be
almost the same when compiling the Object Dictionary.

\subsection{Classes}

Information inheritance has been widely used in Artificial Intelligence, as an
element of some knowledge representation mechanisms, as well as a limited
reasoning device. Our language defines classes as bundles of feature-value
sets that can be inherited by any particular entry defined to be a member of a
class.  The entries belonging to a particular class inherit all the
feature-value pairs present in their parent class. Inheritance is overridden
for those feature-value pairs explicitly stated in the entry. This mechanism
(default inheritance) provides a convenient and natural way to express
regularities and exceptions. Classes can be members of other classes if
desired, so it is possible to build complex inheritance hierarchies that group
and optimize the information organization. Multiple inheritance is also
allowed, so a priority schema has been adopted to avoid conflicts.

A class definition is a label (EN) and a feature structure (ES) attached to
it. If the class defined is a member of a set of other classes, these are
listed in parenthesis after the EN. This is true for entries in other
sections also (words, morphemes and lemmas).

Allomorphy rules are usually stated in a class definition. Rule invocation,
however, is made when a particular child entry from that class is
processed. The EN of such entry acts as the argument for the rule.  As an
example we show partially one of the verbal models we are using:
\begin{quote}\begin{verbatim}
#CLASSES

MV                       % General verbal model
concat = vl
alo 1 stem = $rv0        % Rule invocation to get the value

MV8c (MV)       % Verbal model 8c
alo 1 stt = 0 14 15 21 22 23 24 25 26 31 32 34 35 \
            41 42 43 44 45 46 71 72 73 74 75 76 85 99
alo 1 sut = reg
alo 2 stem = $rv8c
alo 2 stt = 11 12 13 16 33 36 51 52 53 54 55 56 \
            61 62 63 64 65 66 82 90
alo 2 sut = reg
\end{verbatim}\end{quote}

\subsection{Lemmas}

A {\em lemma\/} is a grouping of related entries that share common
information. Each lemma will be expanded to different entries when the Object
Dictionary is compiled. For our purposes a lemma groups the allomorphs needed
to build all the inflected forms, not all the possible surface realizations
--for verbs, where around 53 word forms are possible\footnote{Considering
simple forms only, and excluding the archaic subjunctive future forms and any
clitic agglutination.}, a maximum of eight allomorphs are encoded.

For regular inflection, entries can be very short if the inheritance
mechanism is used. For very irregular lemmas, the entry is usually longer,
because all the information must be provided inside.  We show a very simple
example, that extensively uses the inheritance mechanism:
\begin{quote}\begin{verbatim}
#LEXEMES

pedir (MV8c C3)
\end{verbatim}\end{quote}

\subsection{Allomorphy Rules}

Allomorphy rules are declared in a separate section, and are designed to
build particular allomorphs for a given {\em lemma\/} entry. Rule invocation
is usually done in a class definition, although it can be done in a particular
entry in the lemmas section. The examples that illustrated above the classes
section had two rule invocations.  These always happen as the value for a
particular feature, and the value returned by the rule is assigned to such
feature. Rule invocation is made by name, preceding it with the special
character \$. A special identity allomorphy rule invocation is provided as
the token \$\$. As it was said before, any allomorphy rule invocation takes
as argument the relevant EN for the entry under consideration, and if
invocation takes place in a class definition the argument is the EN of the
entry that inherits that feature.

Rule application is a pattern-matching process. The argument of the rule is
matched sequentially against the {\em left hand side\/} of each production in
the rule. When a match succeds, the relevant right hand side is returned. If
there is no successful match, the rule fails and neither value is returned nor
assigned to the feature that invoqued the rule.  Patterns in the left hand
side of the rule are a sequence of characters and variables --that represent a
regular-expression pattern declared in a rule header. When the argument is
succesfully matched against the left had side of the current production,
variable instantiation takes place. If the right hand side of the production
contains that variable, its instantiated value is used to compute the returned
value.

Formally, a rule contains a name, followed by local variable declarations and
one or more productions, whose left and right sides are separated by the
special token $-$$>$. Variable declarations are assignments --enclosed in
brackets-- of regular expressions to the variable identifier (some single
alphabetic character). Regular expressions are the standard ones of the UNIX
operating system, so they will not be discussed here. Variable invocation in
the productions are preceded by the special character \$.

The following example shows the rule that appeared in an example above. It
computes an allomorph from an infinitive form when it finishes in {\tt -e{\sc
C}ir}, being {\sc C} any consonant, by changing {\tt e} to {\tt i} and
deleting the {\tt ir} ending.  This example has only one production.
\begin{quote}\begin{verbatim}
#ALO-RULES

rv8c
{X = .+}                         % Any non empty character sequence
{C = [bcdfghjklmn'npqrstvwxyz]}  % Any single consonant
$Xe$Cir -> $Xi$C                 % for example: pedir -> pid
\end{verbatim}\end{quote}

\subsection{Type Checking}

This particular section has been designed to providee some kind of type
checking.  Open and closed features have to be declared here. For closed
features all the possible values have to be declared too. We will show an
example:
\begin{quote}\begin{verbatim}
#DATA-DICT

stem =
pers = 1 2 3
agr = @(gen num) @(num pers)
\end{verbatim}\end{quote}

{\tt stem} is an open feature that can take any atomic value (including
character strings), while {\tt pers} is a closed one, and its legal values are
only {\tt 1, 2 {\rm or} 3}. {\tt agr} is a closed feature that can take only a
feature structure as its value, and some restrictions are declared over its
possible feature components: {\tt gender} and {\tt number}, or {\tt number}
and {\tt person}\footnote{It is possible that only one of the features
appears, but not {\tt gen} and {\tt pers} at the same time.}.

This section is of special interest for consistency checking over the whole
source lexical base, for detecting misspellings of feature names and values,
and as reference for lexicographer editors. It will be used also by some tools
to improve the efficiency of the deliverable products, as closed feature
values can be coded since they form a finite set.

\subsection{Object Dictionary generation}

In this section of the Source Lexical Base, rules are given for building the
Object Dictionary. Each rule is a sequence of tree manipulating operators that
can be used to modify the tree structure, filter out or add some branches to
it.

The section specifies a set of rules for each of the sections  containing
lexical entries (lemmas, words and morphemes). From the point of view of
Object Dictionary construction all these three sections are equivalent, and it
is because of these rules that they behave differently.

This section is split in three subsections, each one headed by one of the
labels {\tt LEXEMES, MORPHEMES {\rm or} WORDS}. The rules in each subsection
will be applied to the entries defined in the relevant section of the Source
Lexical Base. For each rule succesfully applied, a new entry will be generated
in the Object Dictionary.

Each rule consist of a sequence of equations. The left hand side refers to the
entry generated in the Object Dictionary and the right hand side to the entry
under consideration in the Source Lexical Base. The special tokens \$\$ and @
refer to the EN and to the ES respectively. All rules must have a value
assigned to \$\$ and to @, and the rule is successful if an effective value is
assigned to \$\$ at runtime (assigned values might not exist).

Tree branches can be accessed by path from @, and assignments to non-existing
branches are considered tree augmenting. Deleting a branch is done by
specifying an incomplete copy: in the right hand side of an equation, after a
subtree specification, a sequence of paths to eliminate is written into
parenthesis, preceding each one with a minus ($-$) sign. Rules are
invoqued sequentially, and non-monotonically: an equation can override a
value assigned by previous equations.

For each possible allomorph that can be included in a lemma entry, a rule
should be included in this section. We have not considered iteration to
enhance this tree manipulating language to cope with an indeterminate number
of allomorphs, because we have always found a manageable number of them.

The example shows that the entries in the words and morphemes sections are
just copied. For the {\em lemmas\/} the same set of operations is repeated for
each of the possible allomorphs (we show just the relevant rule for the third
one). The allomorph is converted to the EN and the older EN becomes the
feature {\tt lex} of the target ES. Some deleting is also done:
\newpage
\begin{quote}\begin{verbatim}
#DICT-RULES

WORDS

@  = @
$$ = $$

MORPHEMES

@  = @
$$ = $$

LEXEMES

$$ = @ alo 3 stem
@  = @ alo 3 (- stem)
@  = @ (- alo  - aux)
@ lex = $$
\end{verbatim}\end{quote}

\section{Conclusions}

The need for the development of our own lexical framework arose from the fact
that other existing formalisms presented some inadequacies in order to cover
the goals we stated at the first section of this paper. Verbosity or strong
ideological commitment to particular linguistic theories were the main
drawbacks found in approaches like \cite{alep} or \cite{RitchieEtAl1987}. But
some good ideas were extracted from these and others, like
\cite{RussellEtAl1991}, \cite{Briscoe1991} or \cite{Haus91}. Although we tried
to minimize ideological commitment, some definitive decisions had to be made,
like abandoning the two-level morphology approach found almost in every
approach, due mostly to the morphological model that we adopted. This has beed
proved useful, since the formalism has been successfully applied to account for
the morphology of the Spanish language with an extensive coverage.

But this framework would be useless if it were computationally intractable. A
set of software tools has beed designed around it, setting--up the basis of
our lexical platform. Extensive work is being carried out at our site
to develop a loosely coupled, highly modular environment that allows to
integrate this set of tools. Among them we will cite efficient
dictionary access libraries, conversion tools between source and object
formats (this includes {\em regexp} rule interpretation, multiple inheritance
management, etc.), and morphological analyser and generator.


\begin{thebibliography}{}

\bibitem[ALEP, 1993]{alep}
P--E International
\newblock (1993).
\newblock ALEP--1 version 1.1 User's Guide. Preliminary Documentation.
\newblock {\em Comission of of the European Communities.}


\bibitem[Briscoe, 1991]{Briscoe1991}
Briscoe, Ted.
\newblock (1991).
\newblock Lexical Issues in Natural Language Processing.
\newblock {\em In Natural Language and Speech, pp. 39--68. Esprit Basic
Research Series Symposium. Springer Verlag.}

\bibitem[Hausser, 1991]{Haus91}
Hausser, Roland,
\newblock (1991).
\newblock Principles of Computational Morphology.
\newblock {\em Unreferenced report distributed electronically.}

\bibitem[Go\~{n}i et. al., 1993]{Go93}
Go\~{n}i, Jos\'{e} M.; Gonz\'{a}lez, Jos\'{e} C. and L\'{o}pez, Jes\'{u}s.
\newblock (1993).
\newblock Espe\-ci\-fi\-ca\-ci\'{o}n del For\-ma\-lis\-mo L\'{e}xico para
ARIES.
\newblock {\em UPM-DIT-GSI Internal Report.}

\bibitem[Moreno, 1991]{Mor91}
Moreno, Antonio.
\newblock (1991).
\newblock Un Modelo Computacional Basado en la Uni\-fi\-ca\-ci\'{o}n para el
        An\'{a}lisis y la Generaci\'{o}n de la Morfolog\'{\i}a
        del Espa\~{n}ol.
\newblock {\em PhD. Thesis.}

\bibitem[Moreno et. al., 1994]{MorGo94}
Moreno, Antonio; Go\~{n}i, Jos\'{e} M.; Gonz\'{a}lez, Jos\'{e} C. and
Olmedo, Cristina.
\newblock (1994).
\newblock A Morphological Model and Processor for Spanish.
\newblock {\em UPM-DIT-GSI Internal Report.}


\bibitem[Ritchie et. al., 1987]{RitchieEtAl1987}
Ritchie, Graeme D.; Pulman, Stephen G.; Black, Alan W. and Russell, Graham J.
\newblock (1987).
\newblock A Computational Framework for Lexical Description.
\newblock {\em Computational Linguistics, vol. 13, n. 3--4, pp.290--307.}

\bibitem[Russell et. al., 1991]{RussellEtAl1991}
Russell, Graham J.; Carroll, John  and Warwick-Armstrong, Susan.
\newblock (1991).
\newblock Multiple Default Inheritance in a Unification-Based Lexicon.
\newblock {\em In Proceedings of the 29th Annual Meeting of the Association
for Computational Linguistics, pp.215--221.}

\bibitem[Shieber, 1986]{Shi86}
Shieber, Stuart M.
\newblock (1986).
\newblock An Introduction to Unification-Based Approaches to Grammar.
\newblock {\em CSLI Lecture Notes. Center for the Study of Language and
	Information. Stanford University.}

\end{thebibliography}
\end{document}